%% file: paper.tex
\documentclass{article}

\usepackage[preprint]{neurips_2026}

\usepackage[utf8]{inputenc}
\usepackage[T1]{fontenc}
\usepackage{amsmath,amssymb,mathtools}
\usepackage{booktabs}
\usepackage{float}
\usepackage{algorithm}
\usepackage{graphicx}
\usepackage{microtype}
\usepackage{xcolor}
\usepackage{url}
\usepackage{hyperref}

\newcommand{\R}{\mathbb{R}}
\newcommand{\Bstar}{\mathcal{B}^{\star}}
\newcommand{\TopK}{\operatorname{TopKSum}}

\newcommand{\TopKset}{\operatorname{Top\text{-}\!K}}

\title{Exact Quantile Balancing and Load-Error Injection for Mixture-of-Experts}

\author{%
  Pit Neitemeier\thanks{Equal contribution}\textsuperscript{\phantom{*},}\thanks{Correspondence to \href{mailto:pit.neitemeier@aleph-alpha-research.com}{\texttt{pit.neitemeier@aleph-alpha-research.com}}.}
  \quad
  Jiaze Li\footnotemark[1]
  \quad
  Alessio Serra
  \quad
  Philipp Scholl
  \quad
  Sohir Maskey \\
  Aleph Alpha
}

\hypersetup{
  pdftitle={Exact Quantile Balancing and Load-Error Injection for Mixture-of-Experts},
  pdfauthor={Pit Neitemeier, Jiaze Li, Alessio Serra, Philipp Scholl, Sohir Maskey}
}

\begin{document}
\maketitle

\begin{abstract}
Mixture-of-Experts (MoE) training requires global load balance to prevent expert under-utilization and local balance for efficient expert-parallel execution. Existing distributed Quantile Balancing (QB) uses shard-dependent or approximate global quantiles, while token-independent expert biases cannot ensure microbatch-level balance. We introduce \emph{Exact Quantile Balancing} (EQB), which computes exact global-batch BF16 quantiles with negligible communication, and \emph{Load-Error Injection} (LEI), which injects local load errors directly into router-score gradients. On 7.5B-parameter MoEs trained for up to 500B tokens, EQB improves global balance and downstream performance over naive QB, while LEI improves local balance and outperforms the GShard loss at comparable quality.
\end{abstract}

\section{Introduction}

Sparse MoE models increase parameter capacity without proportionally increasing per-token compute by routing each token to only few experts~\citep{shazeer2017outrageously,fedus2022switch}. Realizing this benefit requires balancing expert load at two scales.
\emph{Global balance}, measured over all tokens in an optimizer step, prevents routing from persistently concentrating on a small subset of experts. \emph{Local balance} within an expert-parallel (EP) microbatch reduces load skew, improving dispatch and expert-compute efficiency. These objectives are distinct: imbalances across local microbatches can cancel when aggregated over an optimizer step, so global balance does not imply local balance.

Quantile Balancing (QB)~\citep{su2026qb} controls global load through token-independent expert biases computed from routing-margin quantiles, requiring neither an auxiliary loss nor a controller learning rate. In distributed training, however, the required global-batch quantile is not available on any individual data-parallel rank. \citet{su2026qb, marin2026qb} average rank-local quantiles, yielding a statistic that depends on how the batch is partitioned. \citet{kimi2026k3} instead aggregate a global histogram, making the estimate partition invariant but limiting its accuracy to the chosen histogram resolution. We introduce \emph{Exact Quantile Balancing} (EQB), which applies two-pass BF16 radix selection to recover the exact global-batch empirical quantile using two 256-bin all-reduces per layer.

Exact global balancing does not solve local imbalance. A single expert bias is shared across microbatches and cannot adapt to their routing distributions. The GShard auxiliary loss~\citep{lepikhin2021gshard} provides a local training signal, but propagates the load error through normalized routing probabilities, coupling the correction across experts and using a soft surrogate that can change without changing hard top-$K$ assignments. We introduce \emph{Load-Error Injection} (LEI), which instead passes each expert's local load error directly to its router-score gradients. Across our experiments, EQB improves global balance over rank-averaged QB, while LEI improves local balance over the GShard loss at comparable model quality and can improve local balance and average accuracy together. Together, EQB and LEI provide complementary control of global and local MoE load balance.

\section{Preliminaries: MoE Routing and Load Balancing}
\label{sec:qb}

For each token representation $x_t\in\R^d$, an MoE layer contains $E$ routed experts and selects $K$ experts per token. The router produces logits $z_t=W_{\mathrm r}x_t\in\R^E$ and selects $\mathcal I_t=\TopKset(z_t+b)$, where $b \in \mathbb{R}^E$ is the \emph{expert bias} used for load balancing~\citep{wang2024auxiliary}. The bias affects only expert selection; selected experts are weighted by the unbiased sigmoid scores $s_{t, e}=\operatorname{sigmoid}(z_{t, e})$:
\begin{equation}
  y_t=\sum_{e\in\mathcal I_t}s_{t, e}\operatorname{FFN}_e(x_t).
  \label{eq:moe-router}
\end{equation}

For any token batch $\mathcal B$, define the load of expert $e$ as $f_e(\mathcal B)=\sum_{t\in\mathcal B}\mathbf 1[e\in\mathcal I_t]$, with uniform target $\bar f(\mathcal B)=K|\mathcal B|/E$. The \emph{global batch} $\mathcal B_{\mathrm g}$ contains all tokens in one optimizer step, whereas a \emph{local batch} $\mathcal B_{\mathrm l}$ contains the tokens in one rank-local expert-parallel microbatch.

\paragraph{Quantile Balancing.}
Quantile Balancing (QB)~\citep{su2026qb} computes the expert bias directly from the distribution of routing margins. For optimizer step $q$ and token $t$, let $\tau_t$ denote the $(K+1)$st largest entry of $z_t+b^{(q)}$. The value $\tau_t$ denotes the cutoff that has to be exceeded for an expert to enter the top-$K$ set of the token $t$. Since a balanced expert should be selected for a fraction $K/E$ of tokens, QB chooses the corresponding quantile over the margins $\tau_t-z_{t, e}$:
\begin{equation}
  \widehat b_e^{(q+1)}
  =\operatorname{quantile}_{K/E}\{\tau_t-z_{t, e}:t\in \mathcal B_{\mathrm g}\},
  \qquad
  b^{(q+1)}=\widehat b^{(q+1)}-\operatorname{mean}(\widehat b^{(q+1)})\mathbf 1.
  \label{eq:qb}
\end{equation}
The centering removes a common offset that does not affect top-$K$ selection. Statistics from optimizer step $q$ therefore determine the bias $b^{(q+1)}$ used in the next step. QB uses neither an auxiliary loss nor a controller learning rate. Appendix~\ref{app:qb-derivation} derives this update from the balanced-assignment objective.

In distributed training, the global batch is partitioned across data-parallel ranks, so no individual rank observes the global-batch quantile required by QB.
~\citet{marin2026qb} approximate it by computing a quantile on each shard $\mathcal B_{\mathrm g,p}$ and averaging across ranks.
The target statistic, however, is the quantile $b^{\mathrm{global}}_e$ over the entire global batch from Equation~\ref{eq:qb}.
As quantiles do not commute with averaging, $b^{\mathrm{avg}}_e$ generally differs from $b^{\mathrm{global}}_e$ and depends on how the batch is partitioned. \citet{kimi2026k3} instead aggregate histogram counts of the routing margins across ranks, yielding a partition-invariant estimate of the global quantile with communication independent of the number of tokens. However, the estimate remains approximate, as its resolution is limited by the histogram bin width.

\paragraph{Auxiliary load balancing.}
A token-independent bias can correct systematic expert imbalance but cannot adapt separately to different local microbatches. Load balancing can therefore also be imposed through the router gradients. A common choice is the GShard auxiliary loss~\citep{lepikhin2021gshard}. For a local batch $\mathcal B_{\mathrm l}$, define the \emph{hard-load fraction}
$F_e=f_e(\mathcal B_{\mathrm l})/(K|\mathcal B_{\mathrm l}|)$,
the normalized routing probability
$p_{t, e}=s_{t, e}/\sum_j s_{t, j}$,
and its batch mean
$P_e^{\mathrm G}=|\mathcal B_{\mathrm l}|^{-1}\sum_t p_{t, e}$.
The loss
\begin{equation}
\mathcal L_{\mathrm{aux}}
=\alpha\sum_e F_eP_e^{\mathrm G},
\label{eq}
\end{equation}
with $F_e$ treated as stop-gradient, encourages the router to reduce observed load imbalance. Such gradient-based control is complementary to bias-based methods: for example, \citet{deepseekaiv3} combine an expert-bias controller with a sequence-wise auxiliary loss.

\section{Exact Quantile Balancing}
\label{sec:eqb}

The goal of this section is to introduce a distributed, exact quantile balancing methodology. The QB update requires knowing the $r$-th smallest routing margin for each expert, where $r=TK/E$ for a global batch of $T$ tokens. A direct distributed implementation could gather all token-level margins across data-parallel ranks, but its communication grows with $T$. Alternatively, following the idea of \citet{kimi2026k3}, one could utilize an exact histogram over BF16 values, however, this would require up to $2^{16}$ bins per expert. 

\paragraph{Distributed radix selection.} We introduce EQB by adapting radix selection~\citep{alabi2012fast,radik2024}, which locates an order statistic by successively histogramming groups of key bits and retaining the target bin. Unlike radix top-\(k\), EQB only requires the resulting pivot.

We require one global order statistic per expert from margins sharded across data-parallel ranks. We exploit the 16-bit BF16 representation and use an 8-bit radix, reducing selection to exactly two passes over a high and a low byte. Each rank forms 256-bin counts locally, and an all-reduce locates the target bin globally. EQB never materializes a smaller candidate array: the second pass simply rescans local margins whose high byte matches the selected bin. Thus, token-level margins remain local while the exact global-batch quantile is recovered with communication independent of $T$.

In the coarse pass, each rank counts how many margins fall into each of the 256 high-byte ranges and all-reduces the counts. Their cumulative sum identifies the high byte $u_e$ containing the $r$-th smallest margin and its rank $\rho_e$ within that range. The fine pass then counts the low bytes only for margins with high byte $u_e$. Its cumulative counts identify the low byte $v_e$ at rank $\rho_e$. Because the encoding preserves BF16 order, $(u_e,v_e)$ recovers the exact empirical quantile. Moreover, because both passes aggregate global counts, the result is invariant to how tokens are partitioned across data-parallel ranks.

\paragraph{Communication.} EQB communicates $2E\cdot256$ int32 counts per layer, independent of the number of tokens. For $E=256$, this corresponds to $0.5\,\mathrm{MiB}$ per layer. In our experimental setup with 48 MoE layers (Section~\ref{sec:experiments}), this amounts to $24\,\mathrm{MiB}$ per optimizer step. This is $128\times$ smaller than all-reducing a full $2^{16}$-bin BF16 histogram while returning the same exact empirical quantile. Compared with the histogram approximation of \citet{kimi2026k3}, EQB therefore removes the histogram-resolution choice while retaining token-count-independent communication. Algorithm~\ref{alg:eqb} in Appendix~\ref{app:exact_qb} gives the full distributed procedure.

\section{Load-Error Injection}
\label{sec:lei}

For a local batch $\mathcal B_{\mathrm l}$, let $F_e=f_e(\mathcal B_{\mathrm l})/(K|\mathcal B_{\mathrm l}|)$ denote the hard-load fraction defined in Section~\ref{sec:qb}, and let $Q_e=1/E$ be the uniform target. A natural load-balancing objective is
\begin{equation}
\mathcal L_{\mathrm{bal}}=\tfrac12\lVert F-Q\rVert_2^2.
\end{equation}
However, $F$ is determined by the discrete top-$K$ assignment and therefore has zero gradient almost everywhere. To obtain a useful backward signal, consider a differentiable surrogate $P(s)$ and define
\begin{equation}
    \widetilde F
    =
    P(s)+\operatorname{sg}\!\left(F-P(s)\right),
    \qquad
    \mathcal L_{\mathrm{bal}}^{\mathrm{STE}}
    =
    \frac12\lVert \widetilde F-Q\rVert_2^2,
\end{equation}
where $\operatorname{sg}$ denotes stop-gradient. In the forward pass, $\widetilde F=F$. In the backward pass,
\begin{equation}
\label{eq:aux-loss-ste}
    \nabla_s\mathcal L_{\mathrm{bal}}^{\mathrm{STE}}
    =
    J_P(s)^\top(F-Q).
\end{equation}
Thus, the observed hard-load error $F-Q$ determines the desired correction, while the choice of surrogate $P$ determines how that correction is propagated to the router scores.

\paragraph{GShard as a surrogate gradient.}
The GShard loss can be viewed as using mean normalized routing probabilities \(P^{\mathrm G}\) as the STE surrogate. Its probability-dependent Jacobian couples experts and can change without changing the hard top-\(K\) assignments. Appendix~\ref{app:lei-derivation} gives the full derivation.

\paragraph{Load-Error Injection}
To avoid utilizing the normalized routing probabilities $P^G$, which induces the drawbacks outlined above, we introduce \emph{Load-Error Injection} (LEI). LEI applies the mean unnormalized router score 
\begin{equation}
    P_e^{\mathrm L}=\frac{1}{|\mathcal B_{\mathrm l}|}\sum_{t\in\mathcal B_{\mathrm l}}s_{t, e}
\end{equation}
as the STE surrogate. Its Jacobian is constant and diagonal, 
\begin{equation}
\frac{\partial P_e^{\mathrm L}}{\partial s_{t,e'}}
=
\frac{1}{|\mathcal B_{\mathrm l}|}\mathbf 1[e=e'],
\end{equation}
so the load error of expert $e$ affects only that expert's router scores and is applied equally to every token in the local batch. Defining the relative load error $\rho_e=f_e(\mathcal B_{\mathrm l})/\bar f(\mathcal B_{\mathrm l})-1=E(F_e-Q_e)$ LEI therefore applies the simple gradient correction
\begin{equation}
  \operatorname{grad}_{s_{t, e}} \,=\, \operatorname{grad}_{s_{t, e}} \,+\, \eta\,\rho_e,
  \quad t\in\mathcal B_{\mathrm l}.
  \label{eq:lei}
\end{equation}
where $\eta>0$ controls the correction strength. Thus, overloaded experts receive a positive score gradient and underloaded experts a negative one. In contrast to GShard, LEI does not transform the load error through normalized routing probabilities or couple it to the residuals of other experts. Appendix~\ref{app:lei-derivation} provides the complete derivation.

\paragraph{Stabilizing the correction.}
In early ablations, unbounded auxiliary score gradients caused attention logit instability for both the GShard loss and LEI (Figure~\ref{fig:lei-traj}). We therefore bound LEI's injected residual without changing its STE direction, using one shared scale across experts:
\begin{equation}
  \widetilde\rho=\rho\,\frac{\tanh(\xi)}{\xi},
  \qquad \xi=\frac{\lVert\rho\rVert_\infty}{c},
  \label{eq:lei-normalization}
\end{equation}
with $c>0$ and $\tanh(\xi)/\xi:=1$ at $\xi=0$. This preserves the direction and zero-centering of $\rho$ while bounding the injected gradient. LEI leaves forward routing and mixture weights unchanged.

\section{Experiments}
\label{sec:experiments}
\paragraph{Setup.}
We evaluate on a 7.5B-parameter decoder-only MoE with 0.5B active parameters per token, $E=256$ routed experts, and $K=6$. We train six ablations for 100B tokens and additionally compare EQB with normalized LEI at 500B tokens. Experimental details are given in Appendix~\ref{app:experiments}.

\paragraph{Metrics.}
We measure imbalance by
$\operatorname{MaxVio}(\mathcal B)=\max_e[f_e(\mathcal B)/\bar f(\mathcal B)-1]$~\citep{deepseekaiv3}.
Global MaxVio uses loads aggregated over an optimizer step, whereas Local MaxVio is computed on rank-local expert-parallel microbatches. We report worst-layer means over the final 1{,}000 steps.

\begin{table}[t!]
  \centering
  \caption{
  100B-token ablation. Full benchmark results are in Appendix~\ref{app:results100b}.
  }
  \label{tab:results100b}
  \scriptsize
  \setlength{\tabcolsep}{4pt}
  \input{results_table.tex}
\end{table}

\paragraph{EQB improves on rank-averaged QB.}
As shown in Table~\ref{tab:results100b}, replacing rank-averaged quantiles with the exact global-batch statistic reduces Global MaxVio from $0.92$ to $0.74$ (19.4\%) and Local MaxVio from $5.96$ to $5.38$ (9.8\%), while improving BPB on all six reported benchmarks.

\paragraph{LEI improves local balance.}
At 100B tokens, normalized LEI reduces Global/Local MaxVio to $0.60/3.52$, compared with $0.71/4.71$ for the GShard loss, at comparable model quality (Table~\ref{tab:results100b}). At 500B tokens, LEI reduces Local MaxVio from $5.14$ to $4.15$ while average accuracy increases from $48.25\%$ to $48.56\%$; Global MaxVio increases from $0.44$ to $0.63$ (Table~\ref{tab:results500b}). Full trajectories, gradient-weight sweeps, and per-benchmark results are reported in Appendix~\ref{app:experiments}.

\begin{table}[t]
  \centering
  \caption{500B-token EQB vs.\ normalized LEI. Full benchmark results in Appendix \ref{app:results500b}}
  \label{tab:results500b}
  \small
  \setlength{\tabcolsep}{6pt}
  \input{results_table_500b_accuracy.tex}
\end{table}

\section{Conclusion}

EQB and LEI address complementary scales of MoE load balancing. EQB recovers the exact global-batch quantile with token-count-independent communication, while LEI injects local load residuals directly into router-score gradients. EQB improves on rank-averaged QB, and LEI improves balance over the GShard loss at comparable quality. Bounding the injected residual preserves this improvement without attention-logit instability.

\paragraph{Limitations} Our results use one model scale and seed, without estimating variance. GShard and LEI use separately tuned coefficients because the same coefficient produces very different router-gradient magnitudes. Local MaxVio is a proxy for dispatch cost rather than measured EP throughput. %

\bibliographystyle{plainnat}
\bibliography{references}

\appendix
\section{Related work}

Sparse MoE systems commonly encourage balanced routing with an auxiliary objective. The GShard loss~\citep{lepikhin2021gshard} couples hard expert loads to mean routing probabilities, and Switch Transformer~\citep{fedus2022switch} uses a closely related objective. \citet{wang2024auxiliary} instead introduce an auxiliary-loss-free controller that updates expert-selection biases from observed loads but requires a tuned step size; DeepSeek-V3 later adopts this strategy~\citep{deepseekaiv3}. These approaches motivate our separation between global bias control and local score-side correction.

QB~\citep{su2026qb} derives a bias update from quantiles of the balanced-assignment problem and removes the controller learning rate. Distributed implementations differ in the statistic they aggregate: Su's formulation and MARIN~\citep{marin2026qb} average rank-local quantiles, whereas \citet{kimi2026k3} estimates the global-batch quantile with an all-reduced histogram. EQB targets the same global-batch statistic exactly, while LEI complements the token-independent QB bias with a local router-score update.

\section{Balanced assignment, existence, and routing uniqueness}
\label{app:qb-derivation}

For $T$ tokens, $E$ experts, and $1\le K<E$ top-$K$ routing, assume $r=TK/E\in\mathbb N$. Define the balanced assignment set
\begin{equation}
  \mathcal M=\left\{M\in\{0,1\}^{T\times E}:
  \sum_eM_{t, e}=K,\quad \sum_tM_{t, e}=r\right\}.
  \label{eq:balanced-set}
\end{equation}
This set is nonempty: enumerate the $TK$ assignment slots cyclically over the $E$ experts and give each token $K$ consecutive slots. Since $K\le E$ and $E\mid TK$, each token receives distinct experts and each expert appears exactly $r$ times. Because $\mathcal M$ is finite and nonempty, the balanced oracle value
\begin{equation}
  V^\star=\max_{M\in\mathcal M}\langle z,M\rangle
  \label{eq:oracle}
\end{equation}
is attained.

\paragraph{Bias representation.} For any $M\in\mathcal M$ and expert bias $b\in\R^E$, column balance gives
\begin{align}
  \langle z,M\rangle
  &=\sum_{t,e}M_{t, e}(z_{t, e}+b_e)-r\sum_e b_e \notag\\
  &\le \sum_t\TopK_K(z_t+b)-r\sum_e b_e
  =:\Phi(b).
  \label{eq:dual}
\end{align}
The inequality maximizes each row independently. The common shift of $b$ is irrelevant because $TK=Er$, so fix $H=\{b\in\R^E:\mathbf{1}^\top b=0\}$. The convex hull of $\mathcal M$ is the integral bipartite $b$-matching polytope. LP strong duality and dual attainment therefore give
\begin{equation}
  \min_{b\in H}\Phi(b)=V^\star,
  \qquad \Bstar:=\arg\min_{b\in H}\Phi(b)\ne\varnothing.
  \label{eq:strong-duality}
\end{equation}

More precisely, for any balanced $M$,
\begin{equation}
  \Phi(b)-\langle z,M\rangle
  =\sum_t\left[\TopK_K(z_t+b)
  -\sum_eM_{t, e}(z_{t, e}+b_e)\right]\ge0.
  \label{eq:row-gap}
\end{equation}
Every bracket is nonnegative. If $M$ is an oracle and $b\in\Bstar$, strong duality makes their sum zero, hence every row selected by $M$ is a valid top-$K$ subset of $z_t+b$. Conversely, if a balanced $M$ is row-wise top-$K$ at $b$, Eq.~\eqref{eq:row-gap} is zero, so $M$ and $b$ are primal and dual optimal. Thus every oracle is supported by every optimal bias. At a boundary tie, this means the oracle is one valid realization; an arbitrary row-local tie rule need not return the balanced realization.

\paragraph{Why the QB update is a quantile.} The threshold identity
\begin{equation}
  \TopK_K(x)=\min_{\alpha\in\R}
  \left\{K\alpha+\sum_e[x_e-\alpha]_+\right\}
  \label{eq:topk-threshold}
\end{equation}
expands the reduced dual into
\begin{equation}
  D(\alpha,b)=K\sum_t\alpha_t
  +\sum_{t,e}[z_{t, e}+b_e-\alpha_t]_+-r\sum_e b_e,
  \qquad \Phi(b)=\min_\alpha D(\alpha,b).
  \label{eq:expanded-dual}
\end{equation}
For fixed $b$, a lower-endpoint minimizer is the $(K+1)$st largest adjusted score, $\alpha_t=(z_t+b)_{[K+1]}$. For fixed $\alpha$, the $e$th bias minimizes $\sum_t[b_e-(\alpha_t-z_{t, e})]_+-rb_e$ and is therefore the $r$th smallest value of $\{\alpha_t-z_{t, e}\}_t$. With the quantile convention of Section~\ref{sec:qb}, this is exactly Eq.~\eqref{eq:qb}, up to the common shift removed by centering. Moreover, $\partial_e\Phi(b)\ni f_e(b)-r$, so its subgradient is the hard-load residual. This derivation follows the balanced-assignment view of Kimi K3~\citep{kimi2026k3}; it also shows why a single simultaneous finite-batch update is not in general a complete solution of the coupled dual problem.

\paragraph{Geometry and uniqueness.} Fix an oracle $M^\star$ and, for distinct experts $u,v$, define
\begin{equation}
  C_{uv}=\min_{t:M^\star_{tu}=1,\,M^\star_{tv}=0}
  (z_{tu}-z_{tv}),
  \label{eq:transition-cost}
\end{equation}
with $C_{uv}=+\infty$ when the index set is empty. The preceding row-wise support condition gives the exact characterization
\begin{equation}
  \boxed{\Bstar=\left\{b\in H:
  b_v-b_u\le C_{uv}\ \text{for every finite }C_{uv}\right\}.}
  \label{eq:bias-polytope}
\end{equation}
Indeed, each inequality says that every selected $u$ weakly outranks every unselected $v$ in the corresponding rows; intersecting them is therefore equivalent to supporting $M^\star$. The set is closed and convex. It is also bounded: on $H$, $\TopK_K(b)>0$ for every nonzero $b$, so positive homogeneity and compactness of the unit sphere give $\TopK_K(b)\ge c\lVert b\rVert_\infty$ for some $c>0$. Since
\begin{equation}
  \Phi(b)\ge T\TopK_K(b)-K\sum_t\lVert z_t\rVert_\infty,
\end{equation}
$\Phi$ is coercive on $H$. Hence $\Bstar$ is a nonempty compact polytope.

The oracle itself is generically unique. Since $\mathcal M$ is finite, two different assignments $M,N$ tie only if $\langle z,M-N\rangle=0$. If the score distribution assigns probability zero to each such hyperplane, as any joint density does, a finite union bound shows that no pair ties almost surely. This does \emph{not} make the numerical bias unique. If $M^\star$ is the unique oracle, every $b\in\Bstar$ supports it, and any balanced row-wise top-$K$ realization at $b$ must equal $M^\star$: by Eq.~\eqref{eq:row-gap}, another realization would be another oracle. Thus the solution is unique in routing space while the supporting biases may fill the polytope in Eq.~\eqref{eq:bias-polytope}.

\section{Exact Quantile Balancing Algorithm}
\label{app:exact_qb}

\begin{algorithm}[H]
  \caption{Two-pass EQB.}
  \label{alg:eqb}
  \footnotesize
  \raggedright
  Let $\kappa(\tau_t-z_{t, e})=(h_{t, e},\ell_{t, e})\in\{0,\ldots,255\}^2$ be an order-preserving encoding of each BF16 routing margin into high and low bytes. Assume $r=TK/E\in\mathbb N$.

  \noindent
  \begin{minipage}[t]{0.48\linewidth}
    \raggedright
    \textbf{Coarse pass (high byte)}\\[1pt]
    \begin{tabular}{@{}r@{\;}l@{}}
      \textbf{1.} & $\displaystyle
        H_{p,e,u}
        \gets
        \sum_{t\in\mathcal B_{\mathrm g,p}}
        \mathbf 1[h_{t, e}=u],
        \, u\in\{0,\ldots,255\}$\\[1pt]
      \textbf{2.} & $\displaystyle
        H_{e,u}
        \gets
        \sum_{p=1}^{P} H_{p,e,u}
        \quad\text{(all-reduce)}$\\[1pt]
      \textbf{3.} & $\displaystyle
        \begin{array}{@{}l@{}}
          u_e\gets\min\{u:\textstyle\sum_{j=0}^{u}H_{e,j}\ge r\},\\[-1pt]
          \rho_e\gets r-\textstyle\sum_{j=0}^{u_e-1}H_{e,j}.
        \end{array}$
    \end{tabular}
  \end{minipage}\hfill
  \begin{minipage}[t]{0.48\linewidth}
    \raggedright
    \textbf{Fine pass (low byte)}\\[1pt]
    \begin{tabular}{@{}r@{\;}l@{}}
      \textbf{4.} &
        $\begin{array}{@{}l@{}}
        L_{p,e,v}\gets
        \sum_{t\in\mathcal B_{\mathrm g,p}}
        \mathbf 1[h_{t,e}=u_e,\ell_{t,e}=v],\\[-1pt]
        v\in\{0,\ldots,255\}.
        \end{array}$\\[1pt]
      \textbf{5.} & $\displaystyle
        \begin{array}{@{}l@{}}
          L_{e,v}
          \gets
          \sum_{p=1}^{P}L_{p,e,v}
          \quad\text{(all-reduce)},\\[-1pt]
          v_e
          \gets
          \min\left\{
            v:
            \sum_{j=0}^{v}L_{e,j}\ge\rho_e
          \right\}.
        \end{array}$\\[1pt]
      \textbf{6.} & $\displaystyle
        \begin{array}{@{}l@{}}
          \widehat b_e
          \gets
          \kappa^{-1}(u_e,v_e),\\[-1pt]
          b
          \gets
          \widehat b
          -\operatorname{mean}(\widehat b)\mathbf 1.
        \end{array}$
    \end{tabular}
  \end{minipage}
\end{algorithm}

\section{Straight-through derivation of the GShard loss and LEI}
\label{app:lei-derivation}

We give the full derivation of the common straight-through view used in Section~\ref{sec:lei}. Normalize the observed hard load as $F_e=f_e^b/(TK)$, let $Q_e=1/E$, and consider $\mathcal L_{\mathrm{bal}}=\tfrac12\lVert F-Q\rVert_2^2$. The top-$K$ load $F$ is nondifferentiable. For any differentiable surrogate $P(u)$, define
\begin{equation}
  \widetilde F=P(u)+\operatorname{sg}\!\left(F-P(u)\right),
  \qquad
  \mathcal L_{\mathrm{bal}}^{\mathrm{STE}}
  =\frac12\sum_e(\widetilde F_e-Q_e)^2,
  \label{eq:ste}
\end{equation}
where $\operatorname{sg}$ denotes stop-gradient. In the forward pass, $\widetilde F=F$. In the backward pass, $\nabla_u\widetilde F_e=\nabla_uP_e(u)$, and therefore
\begin{equation}
  \nabla_u\mathcal L_{\mathrm{bal}}^{\mathrm{STE}}
  =\sum_e(F_e-Q_e)\nabla_uP_e(u).
  \label{eq:ste-gradient}
\end{equation}
The gradient is the observed hard-load residual multiplied by a surrogate Jacobian. The surrogate need not be numerically close to $F$; it only needs a Jacobian whose increasing directions track increased expert selection.

For the \emph{GShard loss}, choose the batch-mean normalized routing probability
\begin{equation}
  p_{t, e}=\frac{s_{t, e}}{\sum_j s_{t, j}},
  \qquad
  P_e^{\mathrm G}=\frac1T\sum_t p_{t, e},
\end{equation}
where $s$ is the unbiased router score. Because $\sum_eP_e^{\mathrm G}=1$ and $Q_e=1/E$,
\[
  \sum_eQ_e\nabla_sP_e^{\mathrm G}
  =\frac1E\nabla_s\sum_eP_e^{\mathrm G}=0.
\]
The $Q$ term in Eq.~\eqref{eq:ste-gradient} therefore vanishes. Since $F$ is treated as stop-gradient, the remaining term is
\begin{equation}
  \nabla_s\mathcal L_{\mathrm{bal}}^{\mathrm{STE}}
  =\sum_eF_e\nabla_sP_e^{\mathrm G}
  =\frac{1}{\alpha}\nabla_s\mathcal L_{\mathrm{aux}}.
  \label{eq:gshard-ste-equivalence}
\end{equation}
To make this gradient explicit for each router score, let $d_t=\sum_j s_{t, j}$ and $R=F-Q$. Only the $t$th term in $P_e^{\mathrm G}=T^{-1}\sum_{t'}p_{t',e}$ depends on $s_{t, j}$, so $\partial P_e^{\mathrm G}/\partial s_{t, j}=T^{-1}\partial p_{t,e}/\partial s_{t, j}$. Applying the quotient rule and then summing over experts gives
\begin{equation}
  \begin{aligned}
    \frac{\partial p_{t, e}}{\partial s_{t, j}}
    &=\frac{\mathbf{1}[e=j]-p_{t, e}}{d_t},\\
    \frac{\partial\mathcal L_{\mathrm{bal}}^{\mathrm{STE}}}{\partial s_{t, j}}
    &=\frac1T\sum_eR_e\frac{\partial p_{t,e}}{\partial s_{t, j}}
    =\frac{1}{Td_t}\left(R_j-\sum_e p_{t, e}R_e\right).
  \end{aligned}
  \label{eq:gshard-grad}
\end{equation}
Thus every expert's residual is coupled to all others through the probability-weighted mean, and the magnitude is scaled by the token's current score sum. For a softmax router this is the usual softmax probability Jacobian; for a sigmoid router it is followed by the sigmoid Jacobian when propagated to the router logits.

With an expert bias, this estimator combines the actual load selected using $z+b$ with probabilities computed from $s$ alone, a common implementation choice.\footnote{For example, the \href{https://github.com/NVIDIA-NeMo/Nemotron/blob/e27b37133d652f72dbf631ff8bb487df45f8f7d1/src/nemotron/recipes/ultra3/tiny_model.py}{Nemotron recipe} enables both sequence auxiliary loss and expert bias, while the \href{https://github.com/NVIDIA-NeMo/Nemotron/blob/e27b37133d652f72dbf631ff8bb487df45f8f7d1/usage-cookbook/Nemotron-3-Ultra/RL/checkpoint_compatibility/modeling_nemotron_h.py}{router implementation} applies the bias only to expert selection and computes mixture weights from the original scores.} When $b$ substantially changes the top-$K$ choices, $P^{\mathrm G}$ can be weakly aligned with the observed load; its score-dependent Jacobian can further attenuate the correction.

For \emph{LEI}, choose instead the batch-mean unnormalized score
\begin{equation}
  P_e^{\mathrm L}=\frac1T\sum_t s_{t, e}.
\end{equation}
Its Jacobian is constant and diagonal:
\[
  \frac{\partial P_e^{\mathrm L}}{\partial s_{t, e'}}
  =\frac1T\mathbf 1[e=e'].
\]
Substitution into Eq.~\eqref{eq:ste-gradient} gives
\begin{equation}
  \frac{\partial\mathcal L_{\mathrm{bal}}^{\mathrm{STE}}}
       {\partial s_{t, e}}
  =\frac1T(F_e-Q_e).
\end{equation}
Thus LEI broadcasts the same observed load residual to every token, up to the coefficient and residual rescaling in Eq.~\eqref{eq:lei}. The GShard loss and LEI are therefore both STEs of the same hard-load objective; they differ in the surrogate Jacobian, normalized probability for the GShard loss and constant identity-like raw score for LEI. In both score-side estimators, $b$ determines $F$ but is held fixed during differentiation.

The same view also covers a gradient-based auxiliary-loss-free bias controller. Taking $u=b$ and $P(b)=b$ in Eq.~\eqref{eq:ste-gradient} gives
\begin{equation}
  \nabla_b\mathcal L_{\mathrm{bal}}^{\mathrm{STE}}=F-Q.
\end{equation}
It uses the same identity-like surrogate Jacobian as LEI but acts on the token-independent selection bias rather than on token-dependent router scores. QB replaces a tuned gradient step along this residual with the exact coordinate-wise quantile minimizer of Eq.~\eqref{eq:expanded-dual}.

\section{Experimental Details and Additional Results}
\label{app:experiments}

\subsection{Model, training, and evaluation}
\label{app:experimental-details}
All experiments use a 7.5B-parameter decoder-only model with 0.5B active parameters per token. The model has 50 Transformer layers, with two dense layers followed by 48 MoE layers, and combines sliding-window attention (SWA) with full attention in a 4:1 ratio. Each MoE layer contains $E=256$ routed experts with $K=6$ active experts per token, plus one shared expert. We use sigmoid routing with the expert bias applied to the router logits, a 128k vocabulary, and sequence length 16{,}384.

The six 100B-token ablations use 64 data-parallel ranks, a global batch size of 576 sequences, and a warmup-stable learning-rate schedule. Each run trains for 10{,}342 optimizer steps. To evaluate the balance--quality trade-off over a longer horizon, we additionally train EQB and normalized LEI variants for 500B tokens.

\paragraph{Evaluation.}
Evaluations use the Aleph Alpha Eval Framework~\citep{eval_framework} and include MMLU~\citep{hendrycks2021mmlu}, ARC~\citep{clark2018arc}, HellaSwag~\citep{zellers2019hellaswag}, TriviaQA~\citep{joshi2017triviaqa}, GSM8K~\citep{cobbe2021gsm8k}, HumanEval~\citep{chen2021humaneval}, MBPP~\citep{austin2021mbpp}, and MATH~\citep{hendrycks2021math} following the Minerva evaluation protocol~\citep{lewkowycz2022minerva}.

\paragraph{Balance metrics.}
Using the notation from Section~\ref{sec:qb}, we define
\begin{equation}
\operatorname{MaxVio}(\mathcal B)
=
\max_e
\left[
\frac{f_e(\mathcal B)}{\bar f(\mathcal B)}-1
\right],
\end{equation}
following common MoE usage~\citep{deepseekaiv3}. Global MaxVio is computed from expert loads aggregated over all tokens in an optimizer step. Local MaxVio is computed separately for each rank-local expert-parallel microbatch and then averaged across microbatches and ranks. For both metrics, we take the maximum across MoE layers at each step and report the mean over the final 1{,}000 steps. Thus, a Local MaxVio of $5.96$ corresponds to the most-loaded expert receiving, on average, approximately $6.96\times$ its uniform load within the worst-balanced layer.

We additionally track the maximum log-sum-exp (LSE) of the pre-softmax attention logits across SWA layers, tokens, and heads as an indicator of attention-logit instability.

\subsection{100B-token ablation}
\label{app:results100b}

\begin{figure}[t!]
\centering
\includegraphics[width=\linewidth]{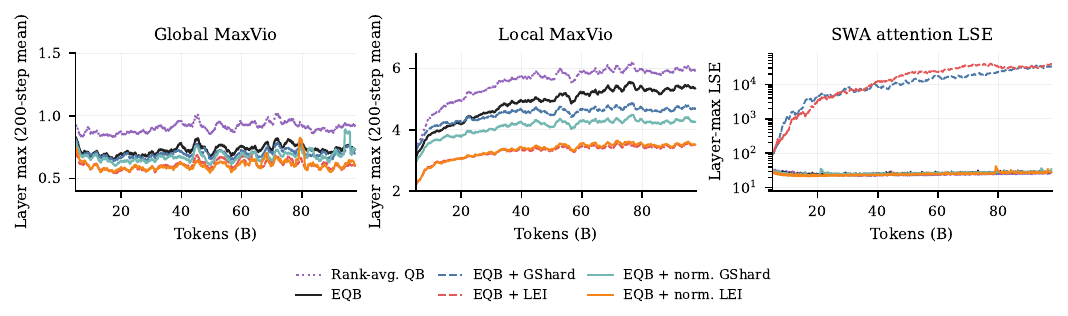}
\caption{100B-token ablation trajectories. LEI reduces both Global and Local MaxVio more strongly than the GShard loss. Without normalization, both score-side balancing methods cause runaway SWA attention LSE; normalization preserves most of the balance improvement while keeping LSE stable.}
\label{fig:lei-traj}
\end{figure}

Figure~\ref{fig:lei-traj} shows the training trajectories corresponding to the endpoint results in Table~\ref{tab:results100b}.
Table~\ref{tab:results100b-extended} reports the complete per-benchmark evaluation.

\begin{table}[t!]
\centering
\caption{Six-run 100B-token ablation, all benchmarks (BPB, lower is better).}
\label{tab:results100b-extended}
\scriptsize
\setlength{\tabcolsep}{3.5pt}
\input{benchmark_table_100b.tex}
\end{table}

\subsection{500B-token results}
\label{app:results500b}

Table~\ref{tab:results500b-sweep} reports the normalized-LEI gradient-weight sweep at 500B tokens.
Table~\ref{tab:results500b-extended} gives the corresponding per-benchmark BPB results.

\begin{table}[t]
\centering
\caption{Normalized-LEI gradient-weight sweep at 500B tokens. Balance metrics are worst-layer means over the final 1{,}000 steps; Avg.\ is the unweighted mean of the reported accuracy metrics.}
\label{tab:results500b-sweep}
\scriptsize
\setlength{\tabcolsep}{1pt}
\input{long_horizon_table.tex}
\end{table}

\begin{table}[t]
\centering
\caption{500B-token setting, all benchmarks (BPB, lower is better).}
\label{tab:results500b-extended}
\scriptsize
\setlength{\tabcolsep}{9pt}
\input{benchmark_table_500b.tex}
\end{table}

\end{document}

%% file: results_table.tex
\begin{tabular}{lrr rrrrrrr}
\toprule
& \multicolumn{2}{c}{MaxVio $\downarrow$}
& \multicolumn{7}{c}{BPB $\downarrow$} \\
\cmidrule(lr){2-3}
\cmidrule(lr){4-10}
Method & Global & Local & MMLU & ARC & HSwag & GSM8K & HumanEval & MBPP & Mean \\
\midrule
Rank-avg. QB
& 0.92 & 5.96
& 0.9007 & 0.7508 & \underline{0.7718} & \underline{0.5260}
& 0.5205 & 0.5993 & 0.6782 \\

EQB
& 0.74 & 5.38
& \textbf{0.8650} & \textbf{0.6876} & \textbf{0.7703} & \textbf{0.5258}
& 0.5154 & \textbf{0.5777} & \textbf{0.6570} \\

EQB + GShard loss
& 0.71 & 4.71
& 0.8932 & 0.7295 & 0.7737 & 0.5400
& 0.5129 & 0.6625 & 0.6853 \\

EQB + LEI
& \textbf{0.60} & \textbf{3.49}
& 0.8897 & 0.7348 & 0.7729 & 0.5409
& \textbf{0.4989} & 0.6177 & 0.6758 \\

EQB + norm. GShard loss
& 0.74 & 4.30
& \underline{0.8853} & 0.7373 & 0.7730 & 0.5333
& \underline{0.5031} & 0.6228 & 0.6758 \\

EQB + norm. LEI
& \underline{0.60} & \underline{3.52}
& 0.8892 & \underline{0.7198} & 0.7729 & 0.5341
& 0.5167 & \underline{0.5992} & \underline{0.6720} \\
\bottomrule
\end{tabular}

%% file: results_table_500b_accuracy.tex
\begin{tabular}{l rr rrrrr}
\toprule
 & \multicolumn{2}{c}{MaxVio $\downarrow$} & \multicolumn{5}{c}{Accuracy (\%) $\uparrow$} \\
\cmidrule(lr){2-3} \cmidrule(lr){4-8}
Method & Global & Local & MMLU & ARC & HSwag & TriviaQA & Avg. \\
\midrule
EQB & \textbf{0.44} & 5.14 & 41.36 & 50.05 & \textbf{63.53} & \textbf{38.06} & 48.25 \\
EQB + norm. LEI & 0.63 & \textbf{4.15} & \textbf{41.38} & \textbf{51.49} & \textbf{63.53} & 37.83 & \textbf{48.56} \\
\bottomrule
\end{tabular}

%% file: benchmark_table_100b.tex
\begin{tabular}{lrrrrrr}
\toprule
Benchmark & Rank-avg. QB & EQB & EQB + GShard loss & EQB + LEI & EQB + norm. GShard loss & EQB + norm. LEI \\
\midrule
\multicolumn{7}{l}{\textit{English general knowledge}} \\
\quad MMLU & 0.9007 & \textbf{0.8650} & 0.8932 & 0.8897 & \underline{0.8853} & 0.8892 \\
\quad ARC & 0.7508 & \textbf{0.6876} & 0.7295 & 0.7348 & 0.7373 & \underline{0.7198} \\
\quad HellaSwag & \underline{0.7718} & \textbf{0.7703} & 0.7737 & 0.7729 & 0.7730 & 0.7729 \\
\quad \textit{aggregate} & 0.8077 & \textbf{0.7743} & 0.7988 & 0.7991 & 0.7985 & \underline{0.7940} \\
\addlinespace
\multicolumn{7}{l}{\textit{English code}} \\
\quad HumanEval & 0.5205 & 0.5154 & 0.5129 & \textbf{0.4989} & \underline{0.5031} & 0.5167 \\
\quad MBPP & 0.5993 & \textbf{0.5777} & 0.6625 & 0.6177 & 0.6228 & \underline{0.5992} \\
\quad \textit{aggregate} & 0.5599 & \textbf{0.5465} & 0.5877 & 0.5583 & 0.5629 & \underline{0.5580} \\
\addlinespace
\multicolumn{7}{l}{\textit{English math}} \\
\quad GSM8K & \underline{0.5260} & \textbf{0.5258} & 0.5400 & 0.5409 & 0.5333 & 0.5341 \\
\addlinespace
\multicolumn{7}{l}{\textit{German general knowledge}} \\
\quad ARC-DE & \underline{0.6616} & \textbf{0.6517} & 0.6666 & 0.6725 & 0.6682 & 0.6633 \\
\quad MMLU-DE & 0.9522 & \textbf{0.9245} & 0.9493 & 0.9559 & \underline{0.9336} & 0.9513 \\
\quad HellaSwag-DE & \textbf{0.6393} & \underline{0.6397} & 0.6438 & 0.6449 & 0.6451 & 0.6398 \\
\quad \textit{aggregate} & 0.7510 & \textbf{0.7386} & 0.7532 & 0.7578 & \underline{0.7490} & 0.7515 \\
\addlinespace
\multicolumn{7}{l}{\textit{German code}} \\
\quad HumanEval-DE & 0.5060 & 0.4869 & 0.5041 & 0.4878 & \textbf{0.4740} & \underline{0.4765} \\
\quad MBPP-DE & 0.6808 & \textbf{0.6505} & 0.6839 & 0.7200 & 0.6835 & \underline{0.6506} \\
\quad \textit{aggregate} & 0.5934 & \underline{0.5687} & 0.5940 & 0.6039 & 0.5788 & \textbf{0.5635} \\
\addlinespace
\multicolumn{7}{l}{\textit{German math}} \\
\quad GSM8K-DE & 0.4691 & 0.4700 & \underline{0.4669} & 0.4698 & 0.4849 & \textbf{0.4612} \\
\quad MATH Minerva-DE & \textbf{0.4923} & \underline{0.4933} & 0.4942 & 0.4958 & 0.4988 & 0.4981 \\
\quad \textit{aggregate} & 0.4807 & 0.4817 & \underline{0.4806} & 0.4828 & 0.4919 & \textbf{0.4797} \\
\bottomrule
\end{tabular}

%% file: long_horizon_table.tex
\begin{tabular}{l rr rrrrr}
\toprule
 & \multicolumn{2}{c}{MaxVio $\downarrow$} & \multicolumn{5}{c}{Accuracy (\%) $\uparrow$} \\
\cmidrule(lr){2-3} \cmidrule(lr){4-8}
Gradient weight $\eta$ & Global & Local & MMLU & ARC & HSwag & TriviaQA & Avg. \\
\midrule
$0$ & \textbf{0.44} & 5.14 & 41.36 & 50.05 & 63.53 & 38.06 & 48.25 \\
$5\times10^{-6}$ & 0.94 & 4.68 & \textbf{43.53} & \textbf{54.05} & \textbf{64.85} & \textbf{39.68} & \textbf{50.53} \\
$2\times10^{-5}$ & 0.63 & \underline{4.15} & \underline{41.38} & \underline{51.49} & 63.53 & 37.83 & \underline{48.56} \\
$8\times10^{-5}$ & \underline{0.48} & \textbf{3.54} & 40.65 & 48.17 & \underline{63.75} & \underline{39.43} & 48.00 \\
\bottomrule
\end{tabular}

%% file: benchmark_table_500b.tex
\begin{tabular}{lrrrr}
\toprule
& \multicolumn{4}{c}{Normalized LEI gradient weight $\eta$} \\
\cmidrule(lr){2-5}
Benchmark & $0$ & $5\times10^{-6}$ & $2\times10^{-5}$ & $8\times10^{-5}$ \\
\midrule
\multicolumn{5}{l}{\textit{English general knowledge}} \\
\quad MMLU & 0.9071 & \textbf{0.8816} & \underline{0.9031} & 0.9125 \\
\quad ARC & 0.8105 & \textbf{0.7582} & \underline{0.7958} & 0.8344 \\
\quad HellaSwag & \textbf{0.7535} & \underline{0.7572} & 0.7595 & 0.7620 \\
\quad \textit{aggregate} & 0.8237 & \textbf{0.7990} & \underline{0.8195} & 0.8363 \\
\addlinespace
\multicolumn{5}{l}{\textit{English code}} \\
\quad HumanEval & 0.4729 & 0.4694 & \underline{0.4422} & \textbf{0.4368} \\
\quad MBPP & 0.5722 & \underline{0.5541} & 0.5931 & \textbf{0.5365} \\
\quad \textit{aggregate} & 0.5225 & \underline{0.5118} & 0.5176 & \textbf{0.4866} \\
\addlinespace
\multicolumn{5}{l}{\textit{English math}} \\
\quad GSM8K & \textbf{0.4689} & \underline{0.4700} & 0.4907 & 0.4745 \\
\quad MATH Minerva & 0.4818 & \underline{0.4796} & \textbf{0.4794} & 0.4881 \\
\quad \textit{aggregate} & \underline{0.4754} & \textbf{0.4748} & 0.4851 & 0.4813 \\
\addlinespace
\multicolumn{5}{l}{\textit{German general knowledge}} \\
\quad ARC-DE & 0.8650 & \textbf{0.8090} & \underline{0.8498} & 0.8629 \\
\quad MMLU-DE & \underline{0.9372} & \textbf{0.9262} & 0.9413 & 0.9506 \\
\quad HellaSwag-DE & 0.6324 & \textbf{0.6236} & \underline{0.6280} & 0.6298 \\
\quad \textit{aggregate} & 0.8115 & \textbf{0.7862} & \underline{0.8064} & 0.8144 \\
\addlinespace
\multicolumn{5}{l}{\textit{German code}} \\
\quad HumanEval-DE & \underline{0.4440} & 0.4503 & 0.4570 & \textbf{0.4176} \\
\quad MBPP-DE & 0.4491 & 0.4483 & \underline{0.4409} & \textbf{0.4352} \\
\quad \textit{aggregate} & \underline{0.4466} & 0.4493 & 0.4490 & \textbf{0.4264} \\
\addlinespace
\multicolumn{5}{l}{\textit{German math}} \\
\quad GSM8K-DE & 0.4234 & \textbf{0.4162} & \underline{0.4212} & 0.4365 \\
\quad MATH Minerva-DE & 0.4007 & \textbf{0.3947} & \underline{0.3955} & 0.4046 \\
\quad \textit{aggregate} & 0.4121 & \textbf{0.4055} & \underline{0.4084} & 0.4205 \\
\bottomrule
\end{tabular}